\documentclass[runningheads, envcountsame, a4paper]{llncs}
\usepackage[pdftex]{graphicx}
\usepackage{epstopdf}
\usepackage{graphicx}
\usepackage{comment}
\usepackage[ruled,vlined]{algorithm2e}
\usepackage{amsmath}
\usepackage{amssymb}
\usepackage{ mathrsfs }
\usepackage{arydshln}
\usepackage{multirow}
\usepackage{wrapfig}
\usepackage{lipsum}
\DeclareMathOperator*{\argmax}{arg\,max}  
\DeclareMathOperator*{\argmin}{arg\,min}  
\graphicspath{ {./images/} }
\SetAlFnt{\footnotesize}
\SetAlCapFnt{\footnotesize}
\begin{document}
%
\title{On Saliency Maps and Adversarial Robustness}
%
%
\author{Puneet Mangla \and
Vedant Singh\and
Vineeth N Balasubramanian}
\authorrunning{P. Mangla et al.}
%
\institute{Department of Computer Science and Engineering \\
IIT Hyderabad, India\\ 
\email{\{cs17btech11029,cs18btech11047,vineethnb\}@iith.ac.in}}
%
\maketitle              
\begin{abstract}
 A very recent trend has emerged to couple the notion of interpretability and adversarial robustness, unlike earlier efforts which solely focused on good interpretations or robustness against adversaries. Works have shown that adversarially trained models exhibit more interpretable saliency maps than their non-robust counterparts, and that this behavior can be quantified by considering the alignment between input image and saliency map. In this work, we provide a different perspective to this coupling, and provide a method, Saliency based Adversarial training (SAT), to use saliency maps to improve adversarial robustness of a model. In particular, we show that using annotations such as bounding boxes and segmentation masks, already provided with a dataset, as weak saliency maps, suffices to improve adversarial robustness with no additional effort to generate the perturbations themselves. Our empirical results on CIFAR-10, CIFAR-100, Tiny ImageNet and Flower-17 datasets consistently corroborate our claim, by showing improved adversarial robustness using our method. We also show how using finer and stronger saliency maps leads to more robust models, and how integrating SAT with existing adversarial training methods, further boosts performance of these existing methods. 

\end{abstract}
\section{Introduction}
\label{intro}
Deep Neural Networks (DNNs) have become vital to solve many tasks across domains including image/text/graph classification and generation, object recognition, segmentation, speech recognition, etc. As the applications of DNNs widen in scope, \textit{robustness} and \textit{interpretability} are two important parameters that define the goodness of a trained DNN model. While on one hand the deep network should be robust to imperceptible perturbations, on the other hand it should be interpretable enough to be trusted when practically used in domains like autonomous navigation or healthcare. Keeping in mind the vulnerability of deep networks to adversarial attacks \cite{szegedy2013intriguing}, efforts have been undertaken to make them more robust to these attacks. Among the proposed methodologies, Adversarial Training (AT) \cite{at@madry,at@goodfellow} has emerged as one of the best defenses wherein networks are trained on adversarial examples to better classify them at test time. On the other hand, in order to generate an interpretable explanation to a network prediction, many methods have been proposed lately, of which guided-backpropagation (GBP)\cite{gbp}, GradCAM++ \cite{grad-cam++}, Integrated Gradients (IG) \cite{integrated-gradients} and fine-grained visualisations (FGVis) \cite{fgvis} are popular to name a few. \par

\begin{figure*}[h]
\centering
\includegraphics[width=\textwidth,height=\textheight,keepaspectratio]{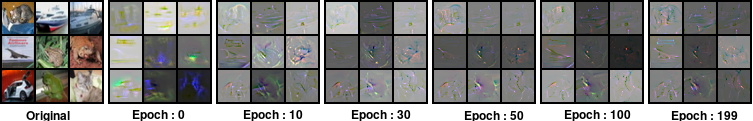}
\caption{Variation of adversarial perturbation with training epochs during 5-step PGD adversarial training of Resnet-34 on CIFAR-10. We observe that adversarial perturbations at later stages of training correspond to class-discriminative regions/pixels.}
\label{fig:pert-variation}
\end{figure*}

Most work so far on interpretability and adversarial robustness have focused on either of them alone. Very recently, over the last few months, there has been a new interest in coupling the two notions and aiming to understand the connection between robustness and interpretability \cite{connection@etmann,imagenet-texture,robustness-odds,interpret-atcnn,attribute-regularization,jacobianAT}. The few existing efforts can be categorized broadly into three kinds: (i) Efforts \cite{interpret-atcnn,robustness-odds} which have shown that explanations generated by more robust models are more interpretable than their non-robust counterparts; (ii) Efforts that have attempted to theoretically analyze the relationship between adversarial robustness and interpretability (for e.g., \cite{connection@etmann} obtains an inequality between these notions which holds to equality in case of linear models, and \cite{relating-explanations} relates the two using a generalized form of hitting set duality); and (iii) Efforts \cite{attribute-regularization,jacobianAT}, more recent, that have attempted to improve robustness by training models with additional objectives that constrain the image explanations to be more interpretable and robust towards attribution attacks. 
All of these efforts are recent, and more work needs to be done to explore the connections better. In this work, we provide a different, yet simple and effective, approach to leverage saliency maps for adversarial robustness. 
We observe that adversarial perturbations correspond to class-discriminative pixels in later stages of training (see Figure \ref{fig:pert-variation}), and exploit this observation to use the saliency map (static) of a given image, whilst training to improve robustness. For datasets where bounding boxes and segmentation masks are provided, we demonstrate that one can exploit these, in lieu of saliency maps, to improve model robustness using our approach. Our work would be closest to the third category of methods described above, and we differentiate from these other recent efforts further in Section \ref{sec_related_work}.


Generally speaking, humans tend to learn new tasks in a robust, generalizable fashion when provided with explanations during their learning phase. For example, a medical student learns about a disease better when provided with explanations of how the disease acts inside the body, rather than just the description of external symptoms. A person may otherwise learn irrelevant relationships without knowledge of underlying explanations. Similarly, we opine that a DNN model that is trained with explanations is less easily fooled by adversarial perturbations. 
Little has been explored in this direction, and we aim to leverage this relationship to provide an efficient methodology for adversarial robustness. We differ from earlier efforts in our primary objective that we use explanations to generate pseudo-adversarial examples instead of using them as a regularization term like \cite{attribute-regularization,jacobianAT} (more in Section \ref{sec_related_work}). In particular, we provide a simple yet effective methodology (as compared to a recent effort such as \cite{relating-explanations}, which is NP-hard) that improves model robustness using saliency maps or already provided bounding boxes or segmentation masks. 
\par

Our key contributions can be summarized as follows: (i) We observe a tangible relationship between a saliency map and adversarial perturbations for a given image, and leverage this observation to propose a new methodology that uses the saliency map of the image to mimic adversarial training; (ii) We show through our empirical studies on widely used datasets: CIFAR-10, CIFAR-100, Tiny ImageNet and Flowers-17, to show the improvement in adversarial robustness through our proposed method; (iii) We show that the improvement becomes more pronounced when a finer and stronger saliency map is used, signifying a strong correlation between saliency maps and adversarial robustness; (iv) When bounding boxes or segmentation masks are already available in a dataset (e.g. Tiny ImageNet or Flowers), we demonstrate that our methodology improves adversarial robustness with no additional cost to generate perturbations for training; (v) Additionally, we show that integrating our method with adversarial training methods, such as PGD \cite{at@madry} and TRADES \cite{trades}, further improves their performance; and (vi) We perform detailed ablation studies to better characterize the efficacy of our proposed methodology. We believe that this work can contribute to opening up a rather new direction to enhance robustness of DNN models.

\section{Related Work}
\label{sec_related_work}
We review earlier efforts related to this work from multiple perspectives, as described below.

\noindent \textbf{Explanation methods:} Various methods have been proposed over the last few years to explain the decisions of a neural network. \textit{Backpropagation-based methods} find the importance of each pixel by backpropagating the class score error to the input image. An improved and popular version of this, known as Guided-Backpropagation \cite{gbp}, only keeps paths that lead to positive influence on the class score, leading to much cleaner-looking explanations. SmoothGrad \cite{smooth-grad}, VarGrad \cite{var-grad} and Integrated gradients \cite{integrated-gradients} refine the explanations by combining/integrating gradients of multiple noisy/interpolated versions of the image. Other backpropagation based methods like DeepLift \cite{deep-lift}, Excitation BackProp \cite{excitation-bp} and Layerwise Relevance Propagation \cite{lrp} generate explanations by utilizing topdown relevance propagation rules. PatternNet and PatternAttribution \cite{patternnet} yields explanations that are theoretically sound for linear models and produce improved explanations for deep networks. CAM \cite{cam}, Grad-CAM \cite{grad-cam}, Grad-CAM++ \cite{grad-cam++} form another variant of generating explanations known as \textit{Activation-based methods}. These methods use linear combinations of activations of convolutional layers, with the weights for these combinations obtained using gradients. \textit{Perturbation  based  methods} generate attribution maps by examining the change in prediction of the  model when the input image is perturbed \cite{pertexp1,pertexp2}. All the aforementioned methods however focus solely on explaining neural network decisions.

\noindent \textbf{Adversarial Attacks and Robustness:} With the advancement of newer adversarial attacks each year \cite{at@goodfellow,at@madry,spatial-adv-exmp,trades}, methods have been proposed to defend against them. Parseval Networks \cite{parseval-nets} train robust networks by constraining the Lipschitz constant of its layers to be smaller than 1. Another category of methods harness the susceptibility of latent layers by performing latent adversarial training (LAT) \cite{LAT}  or using feature denoising \cite{feature-denoising}. Other methods like DefenseGAN \cite{defense-gan} exploit GANs wherein they learn the distribution of unperturbed images and find the closest image for a given test image, to feed the network at inference time. TRADES \cite{trades}, more recently, presents a new defense method that provides a trade-off between adversarial robustness and vanilla accuracy by decomposing prediction error for adversarial examples (robust error) as the sum of natural (classification) error and boundary error, and providing a differentiable upper bound using theory based on classification-calibrated loss. Among the proposed defenses against adversarial attacks, Adversarial Training (AT) \cite{at@madry,at@goodfellow} has remained the most popular and widely used defense, where the network is trained on adversarial examples in order to match the training data distribution with that of adversarial test distribution. More recent efforts in this direction include \cite{free-AT,YOPO}, which aim to reduce adversarial training overhead by recycling gradients and accelerating via the maximal principle respectively. In this work, we focus on adversarial training, considering it still remains the most reliable defense against different attacks.

\noindent \textbf{Interpretability and robustness:} The last few months have seen a few efforts on associating the notions of robustness and interpretability. These efforts can be categorized into three kinds, as introduced briefly earlier: (i) The first kind centers around interpreting how adversarially trained convolutional neural networks (ATCNNs) recognize objects. Recent studies \cite{interpret-atcnn,robustness-odds} have shown that representations learned by ATCNNs are more biased towards image shape than its texture. Also, these ATCNNs tend to evince more interpretable saliency maps corresponding to their prediction than their non-robust equivalents. (ii) In the second kind, inspired by \cite{interpret-atcnn,robustness-odds}, Etmann et al. \cite{connection@etmann} recently quantified this behavior of ATCNNs by considering the alignment between saliency map and the image as the metric for interpretability. They confirmed that for a linear model, the alignment grows strictly with robustness. For non-linear models such as neural networks, they show that the their linearized robustness is loosely bounded by the alignment metric. \cite{relating-explanations}, which is rather new, provides a theoretical connection between adversarial examples and explanations by demonstrating that both are related by a generalized form of hitting set duality. (iii) Encouraged by the work of Etmann \cite{connection@etmann}, a third category of work more recently sought to answer the question: Do robust and interpretable saliency maps imply adversarial robustness ? Recent efforts \cite{explanations-manipulated,fragile-interpretation} have shown that the explanations of neural networks can also be manipulated by adding perturbations to input examples, which, instead of causing mis-classification, result in a different explanation. To tackle this problem, a recent effort, Robust Attribution Regularization (RAR) \cite{attribute-regularization}, aims to train networks by optimizing objectives with a regularization term on explanations in order to achieve robust attributions. 

In contrast to the abovementioned methods, our work explores a different side to the connection between notions of interpretability and robustness. As explained in Section \ref{intro}, we aim to generate adversarial perturbations from a given saliency map to improve robustness while training a neural network. Our work is perhaps closest to RAR \cite{attribute-regularization} in using explanations (saliency maps) during training, we however differ from them in our primary objective and in that we are using explanations to generate adversarial examples, while RAR seeks to attain attributional robustness. In terms of approach, although one could consider the work by Ignatiev et al. \cite{relating-explanations} as related to ours, they had only preliminary results on MNIST, and focused on providing a theoretical connection between adversarial examples and explanations based on logic and constraint programming, which may not scale to large-scale datasets. 
Our work attempts to show that explanations can yield robustness in standard large-scale benchmark datasets in an efficient way. We further hypothesize that, by advancement through our method, we can actually take advantage of weak explanations like bounding boxes and segmentation masks which may otherwise be left unexploited in a dataset, to train adversarially robust models.


\section{Using Saliency Maps for Efficient Adversarial Training}
\label{methodology}
Our method is motivated by the observation that adversarial perturbations at later stages of training correspond to class-discriminative regions/pixels, as shown in Figure \ref{fig:pert-variation}. We begin describing our method with the necessary notations and preliminaries.
\vspace{3pt}

\noindent \textbf{3.1 Notations and Preliminaries.}
We denote a neural network as $\Phi(. \ ;  \ \theta) : \mathbb{R}^d \rightarrow \mathbb{R}^k $, parametrized by weights $\theta$, which takes an input $\textbf{x} \in \mathbb{R}^d $ and outputs a logit, $\Phi^i(\textbf{x})$, for each of $k$ classes, i.e. $i \in \mathcal{C} = \{1,\cdots,k\}$, $\mathcal{C}$ denoting the set of class labels. 
Without loss of generality, considering an image classification setting, we define a saliency map $\textbf{s}$ corresponding to an input sample $\textbf{x}$ as $\textbf{s} \in [0,1]^d$, where the presence of an object of interest in input $\textbf{x}$ lies between 0 and 1. For a trained network $\Phi$, the unnormalized saliency map for an input $\textbf{x}$ can simply be given as:
$\nabla_{\textbf{x}} \Phi^{i^*}(\textbf{x})$, where $i^* = \argmax_i \Phi^i(\textbf{x})$. (In practice, this gradient is at times computed w.r.t feature maps of intermediate layers, but we leave our notations w.r.t $\textbf{x}$ for ease of understanding.)
\vspace{3pt}

\noindent \textit{Projected Gradient Descent (PGD) Attack:} Prior efforts on adversarial attacks \cite{at@goodfellow} include the Fast Gradient Sign Method (FGSM), a $l_{\infty}$-bounded single step attack which calculates an adversary as: $\textbf{x} + \epsilon \hspace{1mm} \text{sign}(\nabla_\textbf{x} \mathcal{L} \ (\Phi(\textbf{x},\theta),y))$.
A more powerful attack however is the popularly used multi-step variant, also called Projected Gradient Descent (PGD), given by:
\begin{align}
\label{eqn_pgd}
    \textbf{x}^0 = \textbf{x};
    \textbf{x}^{t+1} = \Pi_{\textbf{x}+ \textit{N}} \big( \textbf{x}^{t} + \alpha \hspace{1mm} \text{sign}(\nabla_{\textbf{x}} \mathcal{L} \ (\Phi(\textbf{x},\theta),y)) \big)
\end{align}
\noindent where $\alpha$ is the step size, $\Pi$ is the projection function, and $N$ is the space of possible perturbations. We use PGD as the choice of attack in our experiments, and later change to show (in Section \ref{ablations}) how our method performs against other attacks.
\vspace{3pt}

\noindent \textit{Adversarial Training (AT):} AT \cite{at@goodfellow} is generally used to make the models adversarially robust by matching the training distribution with the adversarial test distribution. Essentially, for AT, the optimal parameter $\theta^*$ is given by: 
\begin{equation}
    \theta^* = \argmin_\theta \mathbb{E}_{(\textbf{x},y) \sim D} \ \left[ \ \max_{\delta \in N}\mathcal{L} \ (\Phi(\textbf{x} + \delta,\theta),y) \ \right]
\end{equation}
Here, the inner maximization $\max_{\delta \in N} \mathcal{L} \ (\Phi(\textbf{x} + \delta,\theta),y)$ is calculated using a strong adversarial attack such as PGD.\\

\noindent \textbf{3.2 Saliency-Based Adversarial Training: Motivation.} 
An adversarial perturbation at input $\textbf{x}$ is given by:
\begin{equation}
\arg\inf_{e \ \in \ \mathbb{R}^d} \{\Vert e \Vert: \argmax_i \Phi^i(x+e) \neq \argmax_i \Phi^i(x) \}
\end{equation}
Etmann et al. \cite{connection@etmann} showed that for most multi-class neural networks, especially ones with ReLU or Leaky ReLU activation functions, which we consider in this work\footnote{In all our experiments, we train networks with ReLU activations which ensures that this assumptions is met. We follow the experiments in \cite{connection@etmann} in this regard, to ensure that our model is locally affine in a given data point's neighborhood.}, the network's score function, $\Phi$ is sufficiently locally-linear in relevant neighbourhood of input $\textbf{x}$, i.e.
\begin{equation}
    \Phi^i(\textbf{x} +  e) \approx \Phi^i(\textbf{x}) + e^T \cdot \nabla_\textbf{x}\Phi^i(\textbf{x}) 
\label{eqn_etmann_locally_linear}
\end{equation}
Leveraging this, an adversarial perturbation, $e$, which is intended as a perturbation to input $\textbf{x}$ which results in a change of predicted label, can be modeled as follows (given $i,j \in \mathcal{C}$):
\begin{align}
    & \argmax_i \Phi^i(x+e) \neq \argmax_i \Phi^i(x)\\
    \Longleftrightarrow \qquad & \exists j \neq i^* : \Phi^j(\textbf{x} + e) > \Phi^{i^*}(\textbf{x} + e)\\
    \Longleftrightarrow  \qquad & \exists j \neq i^* : e^T \cdot (\nabla_\textbf{x}\Phi^j(\textbf{x}) - \nabla_\textbf{x}\Phi^{i^*}(\textbf{x})) > \Phi^{i^*}(\textbf{x}) - \Phi^j(\textbf{x})
\end{align}
The third inequality (7) comes from combining inequality (6) and expression (4). The infimum over $\Vert e \Vert$, which provides a minimal perturbation to change the class label, is achieved by choosing $e$ as a multiple of $\nabla_\textbf{x}(\Phi^{j}(\textbf{x})- \Phi^{i^*}(\textbf{x}))$. (In general, note that the LHS of $e^T z > c$ where $c$ is a constant, is maximized by $e = \Vert e \Vert \frac{z}{\Vert z \Vert}$, leading to $\Vert e \Vert \Vert z \Vert > c$. The infimum of $e$ is then achieved by choosing it as a multiple of $z$). 
The direction of adversarial perturbation then becomes:
\begin{equation} 
\label{eqn_pert_direction}
\nabla_\textbf{x}(\Phi^{j}(\textbf{x}) - \Phi^{i^*}(\textbf{x}))
\end{equation}
This perturbation direction depends on two quantities: (i) $\nabla_\textbf{x} \Phi^{i^*}(\textbf{x})$ the saliency map for the true class $i^*$; and (ii) $\nabla_\textbf{x} \Phi^{j}(\textbf{x})$, the saliency map of $\textbf{x}$ for class $j$ for which the infimum of $e$ is attained. We now analyze the direction of adversarial perturbation separately for the binary and multi-class cases.
\vspace{3pt}

\noindent \textit{Binary case:} Let us consider a binary classifier $h: \textbf{x} \rightarrow \{-1,1\}$ given by: $h = \text{sign}(\Phi(\textbf{x},\theta))$, where $\Phi(\textbf{x},\theta))$ represents the logit of the positive class. Let $\Phi'(\textbf{x})$ denotes the logit of negative class. Applying a sigmoid activation function, we get the probability of the positive and negative class as: $P(y=+1|\textbf{x}) = \frac{1}{1 + \exp^{-\Phi(\textbf{x},\theta)}}$ and $P(y=-1|\textbf{x}) = \frac{1}{1 + \exp^{-\Phi'(\textbf{x},\theta)}}$ respectively.
It is simple to see that the probability of the negative class can also be written as:
\[P(y=-1|\textbf{x}) = 1 - P(y=+1|\textbf{x}) = \frac{1}{1 + \exp^{\Phi(\textbf{x},\theta)}}\]
Rather, the corresponding logit score of the negative class is $-\Phi(\textbf{x},\theta))$. In Eqn \ref{eqn_pert_direction} for a binary classifier, one can hence view $\Phi^{j}(\textbf{x}) =  - \Phi^{i^*}(\textbf{x})$, and define the adversarial perturbation direction in Eqn \ref{eqn_pert_direction} simply as $- \nabla_\textbf{x}(\Phi^{i^*}(\textbf{x}))$.
\vspace{3pt}

\noindent \textit{Multi-class case:} We extend a similar argument to the multi-class setting using an approximation. The direction of adversarial perturbation in the multi-class setting as give in Eqn \ref{eqn_pert_direction} would require finding the class $j$ for which the infimum of $\Vert e \Vert$ is attained. This requires computing the quantity in Eqn \ref{eqn_pert_direction} for all classes, and identifying the $j \neq i^*$ for which the quantity attains the least value. This is compute-intensive. To avoid this computational overhead, we rely on $\Phi^{i^*}$ alone (note that $i^*$, the ground truth label, is known to us at training), and simply propose the use of $- \nabla_\textbf{x}(\Phi^{i^*}(\textbf{x}))$ as the direction of perturbation (as in the binary case). We now argue that this approximation is a reasonable one. Considering the multi-class setting as $k$ binary classification problems, for the binary classifier corresponding to the ground truth class $i^*$, we would have:
\[P(y \neq i^*|\textbf{x}) = 1 - P(y = i^*|\textbf{x}) = \frac{1}{1 + \exp^{\Phi^{i^*}(\textbf{x},\theta)}}\]
In other words, the corresponding logit score of the negative class is $-\Phi^{i^*}(\textbf{x},\theta))$. Approximating the direction of the adversarial perturbation as the average of the directions of the perturbations across the $k$ binary classification problems, we would get the direction of the perturbation to be: $\nabla_\textbf{x}(\sum_{j \neq i^*}\Phi^{j}(\textbf{x}) - k \Phi^{i^*}(\textbf{x}))$. Assuming that each of the classes $l \neq i^* \in \mathcal{C}$ is equally likely to be the $j$ that minimizes $\Vert e \Vert$, it is evident that choosing $- \nabla_\textbf{x}(\Phi^{i^*}(\textbf{x}))$ as the direction of perturbation would be the most conservative option in the expected sense. We show through our experiments that this option works reasonably well.


Since we deal with $l_{\infty}$-bounded perturbations in this work, following PGD (Eqn \ref{eqn_pgd}), we use $- \text{sign} \ (\nabla_\textbf{x} \Phi^{i^*}(\textbf{x}))$ instead of $- \nabla_\textbf{x} \Phi^{i^*}(\textbf{x})$ itself as the perturbation direction.
We complete the above discussion by noting that $\nabla_\textbf{x} \Phi^{i^*}(\textbf{x})$ is the saliency map, \textbf{s}, defined at the beginning of this section. 
In other words, the direction of adversarial perturbation can be obtained using a saliency map (as we also show in our experiments). 

\begin{wrapfigure}[23]{l}{0.37\textwidth}
\vspace{-15pt}
\includegraphics[width=0.37\textwidth]{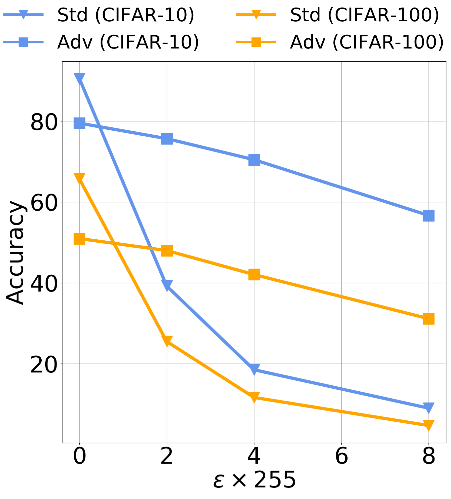}
\vspace{-15pt}
\caption{Adversarial attack accuracy when perturbing input example using the direction of negative saliency on normally (Std.) and adversarially (Adv.) trained models on CIFAR-10 and CIFAR-100. Note that $\epsilon=0$ denotes clean accuracy (no attack), all other models face attacks by saliency maps and have lower accuracies.}
\label{fig:sal-attack}
\end{wrapfigure}
To study the above claim, we conducted experiments to check if the negative of the saliency map corresponding to the ground truth class, i.e. $-\nabla_\textbf{x}\Phi^{i^*}(\textbf{x})$, can indeed be used as a direction to perturb input. We created adversarial examples by perturbing original examples as: $\textbf{x} = \textbf{x} - \epsilon \cdot \text{sign}(\nabla_\textbf{x}\Phi^{i^*}(\textbf{x}))$ (using saliency maps) to attack a model. We attack a Resnet-10 model trained on CIFAR-10 and CIFAR-100 datasets, normally (Std.) and adversarially (Adv.) trained using 5-step PGD attack with 3 different magnitudes of $l_\infty$-norm perturbations, $\epsilon \in \{2/255, 4/255, 8/255\}$. We report our findings in Figure \ref{fig:sal-attack}. The figure shows that, examples generated by perturbing an input in the direction of negative saliency decreases the model's accuracy to a reasonable extent, affecting models both models trained normally and adversarially. We now describe how we leverage this relationship to perform Saliency-based Adversarial Training (SAT).\\

\noindent \textbf{3.3 Saliency-Based Adversarial Training: Algorithm.} 
Algorithm \ref{mimic-AT} summarizes our methodology of using saliency maps for adversarial training. The saliency maps during training are obtained either through annotations provided in a dataset (such as bounding boxes or segmentation masks), or through a pre-trained model which is used only to get saliency maps. More details on obtaining saliency maps is discussed in Section \ref{results}.

We also observed in our studies that when training a network using adversarial training, during the initial phase of training when the weights are not optimal, the perturbations computed by the attack methods are random. But with training, as weights become optimal, they become more class-discriminative. Figure \ref{fig:pert-variation} illustrates this observation, where the perturbation is random in initial phases of training but eventually becomes class-discriminative as the model trains. We exploit this observation to complete our methodology. 
In order to mimic the above behavior of the perturbation over training, we choose the direction of perturbation in a stochastic manner. We choose the $i^{th}$ component $\delta^t$[i] of perturbation $\delta^t$ at time $t$ as:
\begin{equation}
    \label{pert-calc}
    \delta^t[i] =
    \begin{cases}
      \textbf{z}[i], & \text{with probability } \alpha^{t} \\
      -\textbf{s}[i], & \text{with probability } 1-\alpha^{t}
    \end{cases}
 \end{equation}
\noindent where $\textbf{z} \in \{-1,1\}^d$ is sampled randomly, and 0 $< \alpha <$ 1. During initial epochs of training, when $\alpha^t$ is close to 1, $\delta^t$ will be dominated by random values. However, as training proceeds and $\alpha^t$ starts diminishing, $\delta^t$ smoothly transitions to $-\textbf{s}$ and will be influenced by the adversarial character of the saliency map.
\begin{wrapfigure}[21]{l}{0.63\textwidth}
\begin{minipage}{0.63\textwidth}
\vspace{-20pt}
\begin{algorithm}[H]
\SetAlgoLined
\textbf{Input:} Training Dataset $D$, Saliency Maps $S$, Model $\Phi(. \ ; \ \theta)$, SAT hyperparameter $\alpha$, Learning rate $\eta$, Maximum $l_\infty$ perturbation $\epsilon_0$ \\
\textbf{Output:} Optimal parameter $\theta^*$\\
Initialize model parameters as $\theta = \theta^0$. \\
 \For{$t \in \{1, 2, ..., n\}$}
    {
        Sample training data of size $B$ : $\{(\textbf{x}^i, y^i)\}$ from $D$. \\
        Pick out corresponding saliency maps : $\{\textbf{s}^i\}$ from $S$. \\
        Calculate $\delta^{ti}$ for each $\textbf{x}^i$ using Equation \ref{pert-calc}. \\
        Perturb the input examples : $\textbf{x}^i := \textbf{x}^i + \epsilon_0 \cdot \delta^{ti}$. \\
        Perform clipping to keep $\textbf{x}^i$ bounded : $\textbf{x}^i$ := $clip(\textbf{x}^i)$. \\
        Update model parameters : \\ $\theta^t := \theta^{t-1} - \eta \cdot \triangledown_\theta \ \frac{1}{B} \sum_{i=1}^{B} \mathcal{L} \ (\Phi(\textbf{x}^i,\theta^{t-1}),y^i) $
    }
 \caption{Saliency-based Adversarial Training (SAT) Methodology}
 \label{mimic-AT}
\end{algorithm}
\end{minipage}
\end{wrapfigure}
\\
When additional annotations such as bounding boxes or segmentation masks are available in a dataset, our approach considers these as weak saliency maps for the methodology. Usually, bounding boxes or segmentation masks are available as single channel images. To use them in our algorithm, we concatenate them along channel dimension to get the required dimension as image. After this pre-processing, we generate the weak saliency, $\tilde{\textbf{s}}$ from bounding boxes or segmentation masks as: \\

\begin{equation}
    \label{bbox-to-sal}
    \tilde{\textbf{s}}[i] =
    \begin{cases}
      1, & \text{if i\textsuperscript{th} pixel lies inside bbox or seg masks} \\
      -1, & \text{otherwise}
    \end{cases}
 \end{equation}
\noindent
As mentioned earlier, using these bounding boxes or segmentation masks as saliency maps allows us to perform adversarial training with no additional cost to compute perturbations, unlike existing adversarial training methods which can be compute-intensive.

\section{Experiments and Results}
\label{results}
In this section, we present our results using the proposed SAT method (Algorithm \ref{mimic-AT}) on multiple datasets with different variations of saliency maps. We begin with describing the datasets, evaluation criteria and implementation details.\\

\noindent {\large \textbf{4.1 Experimental Setup}}
\vspace{3pt}
\noindent \textbf{Datasets and Evaluation Criteria:} We perform experiments on well-known datasets: CIFAR-10, CIFAR-100 \cite{cifar-datasets}, Tiny ImageNet \cite{tiny-img} and FLOWER-17 \cite{flower}. 
We evaluate the \textit{adversarial robustness} of trained models (model accuracy when perturbations from  following attacks are provided as inputs, as done in all earlier related efforts - also called \textit{adversarial accuracy}) using the popular and widely used PGD attack (described in Sec \ref{methodology}) with 5 steps and 4 different levels of $l_\infty$ perturbation $\epsilon_0 \in \{1/255, 2/255, 3/255, 4/255\}$. We also evaluate the robustness of our models against other attacks: TRADES \cite{trades} as well as uniform noise in Sec \ref{ablations}. We trained all our models for 5 trials and observed minimal variations in the values. We hence report the mean value of our experiments in our tables.
\vspace{3pt}

\noindent \textbf{Baselines:} We compare our method with multiple baseline methods, including those that train adversarially, as well as those that don't train adversarially:
\textit{(i) Original Model:} Model trained normally with no adversarial training; \textit{(ii) PGD-AT:} Model trained adversarially using 5-step PGD attack with max $l_\infty$ perturbation and $\epsilon = 8/255$, as in \cite{at@madry}; \textit{(iii) TRADES-AT:} Model trained adversarially using TRADES attack with max $l_\infty$ perturbation and $\epsilon = 8/255$, as in \cite{trades}; \textit{(iv) Original + Uniform Noise:} Model trained normally with no adversarial training, but training data is perturbed with uniform noise sampled from $[\frac{-8}{255},\frac{8}{255}]$ during training; \textit{(v) PGD + Uniform Noise:} PGD-AT, described in (ii), training data is additionally perturbed with uniform noise sampled from $[\frac{-8}{255},\frac{8}{255}]$ during training; \textit{(vi) TRADES + Uniform Noise:} TRADES-AT, described in (iii), training data is additionally perturbed with uniform noise sampled from $[\frac{-8}{255},\frac{8}{255}]$ during training.
The models are trained for 100 epochs with standard cross-entropy loss, minimized using Adam optimizer (learning rate = $1e-3$).
\vspace{3pt}

\noindent \textbf{Implementation Details:} 
As in Sec \ref{methodology}, our method relies on being provided with a saliency map for the ground truth label of a given image, while training the model. For Tiny ImageNet and FLOWER-17, we simply use the bounding boxes and segmentation masks provided in the dataset (and left unexploited often) as `weak' saliency maps to train the model using the proposed SAT method. 
This thus incurs no additional cost when compared to competing methods that perform adversarial training by generating perturbations (a costly operation when using methods such as PGD). For CIFAR-10 and CIFAR-100, since we do not have such information (bounding boxes/segmentation masks) provided, we obtain saliency maps using teacher networks. In particular, we train two teacher networks: Resnet-10 and Resnet-34 for this purpose. Our final model architecture in all these settings is a Resnet34 too.

\begin{wrapfigure}[9]{l}{0.5\textwidth}
\vspace{-20pt}
\includegraphics[width=0.5\textwidth]{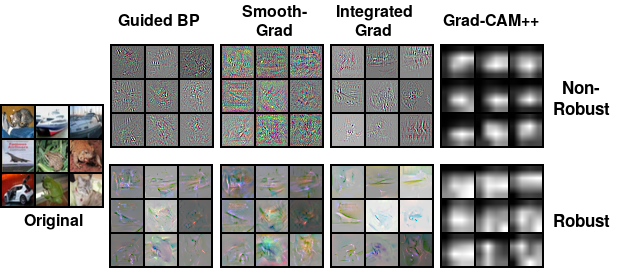}
\vspace{-10pt}
\caption{Saliency maps of non-robust \textit{(top, model trained normally)} and robust \textit{(bottom, model trained adversarially)} variants of Resnet-10 with different explanation methods on CIFAR-10.}
\label{saliency}
\end{wrapfigure}
To go further, we also consider multiple variants of the above choices to study our method more carefully. It is believed that the quality of saliency maps generated by an adversarially trained model is better than its non-robust equivalent \cite{robustness-odds}. Figure \ref{saliency} supports the above claim. We hence train the aforementioned teacher models in two ways: Standard (denoted as \textit{Std} in the results) and Adversarial (denoted as \textit{Adv} in results) (regular PGD-based adversarial training), and use the saliency maps for the ground truth class. The saliency maps themselves are generated in two ways: we use bounding boxes obtained using Grad-CAM++ \cite{grad-cam++} as weak saliency maps, and we use the finer saliency maps generated as is by Guided Back-propagation (GBP) \cite{gbp} to train our student model. We set hyperparameters $\alpha = (0.6)^{\frac{1}{10}}$ and $\epsilon_0 = \frac{8}{255}$, and minimize cross-entropy loss using an Adam optimizer (learning rate = $1e-3$) for 200 epochs.

To complete the study, we also use variants of our method: \textit{PGD-SAT} and \textit{TRADES-SAT}, where we perturb the input randomly with either perturbations calculated by SAT or with PGD/TRADES attacks. 
All these variants are named in a self-explanatory manner in our results: \textit{X | Y | Z}, where  \textit{X} denotes the teacher model (Resnet-10/-34), \textit{Y} denotes the mode of training (Std or Adv), and \textit{Z} denotes the saliency map method (GBP/ GradCAM++) used.
In Sec \ref{ablations}, we also perform additional studies with other saliency methods like Guided-GradCAM++, Smooth-Grad and Integrated-Gradients to study the generalizability of our results on varying quality of saliencies.
\vspace{3pt}

\begin{table}
\begin{center}
\scalebox{0.7}{
\begin{tabular}{c|c|c|c|c|c|c|c|c}
\hline
\textbf{Method}  & \multicolumn{4}{c|}{\textbf{CIFAR-10}} & \multicolumn{4}{c}{\textbf{CIFAR-100}}\\

 & $\epsilon = 1/255$ & $\epsilon = 2/255$  & $\epsilon = 3/255$ & $\epsilon = 4/255$ & $\epsilon = 1/255$ & $\epsilon = 2/255$  & $\epsilon = 3/255$ & $\epsilon = 4/255$ \\
\hline \hline

                        Original & 47.71 & 10.36 & 1.39 & 0.28 & 25.83 & 7.76 & 3.35 & 1.94\\ 
                        Original + Uniform-Noise & 61.23 & 22.85 & 6.34 & 2.56 & 33.15 & 13.50 & 6.01 & 3.22\\
                        \cline{1-9}
                        \textbf{SAT (Weak saliency)} & & & & \\ \cline{1-1}
                          Resnet-10 | Std. | G.CAM++ & 59.61  & 22.12  & 6.16 & 1.77 & 31.98  & 11.93 & 5.48  & 2.89             \\
                          Resnet-10 | Adv. | G.CAM++ & 57.34  & 19.94 & 5.64 & 1.91 & 32.88  & 13.3 & 6.31 & 4.0      \\ 
                          Resnet-34 | Std. | G.CAM++  & 56.75 & 20.62 & 6.02 & 1.74 &  30.95 & 12.68 & 5.65 & 3.25           \\
                         Resnet-34 | Adv. | G.CAM++ & 60.0 & 22.94  & 6.58 & 1.96 & 32.96 & 12.07  & 5.2 & 3.04 \\
                           \cline{1-9} 
                        \textbf{SAT (Fine saliency)} & & & & \\ \cline{1-1}
                         Resnet-10 | Std. | GBP & 10.87  & 0.95  & 0.0  & 0.0 & 20.53  & 7.52  & 3.5  & 2.12              \\
                         Resnet-10 | Adv. | GBP & \textbf{63.33}  & \textbf{26.79} & \textbf{9.62} & \textbf{3.69} & 34.29  & \textbf{14.73} & 6.84 & 4.22        \\ 
                         Resnet-34 | Std. | GBP  & 18.01 & 2.54 & 1.25 & 1.0 & 9.91 & 2.52 & 1.05 & 0.54             \\
                         Resnet-34 | Adv. | GBP & 62.67 & 23.76  & 5.82 & 1.28 & \textbf{34.71} & 14.32  & \textbf{7.08} & \textbf{4.29} \\ 

\hline 
& & & & & & & & \\ 
\hline
                        PGD & 77.89 & 73.1 & 66.96 & 61.18 & 45.75 & 40.13 & 35.41 & 31.01 \\ 
                        PGD + Uniform-Noise & \textbf{82.23} & 74.97 & 65.61 & 55.04 & \textbf{42.67} & 36.10& 30.64 \\ 
                        \cline{1-9}
                        \textbf{PGD-SAT} & & & & \\
                        \cline{1-1}
                        Resnet-10 | Std. | GBP & 79.72 & 73.81 & 67.72 & 61.29 & 46.87 & 41.11 & 35.77 & 30.75   \\
                        Resnet-10 | Adv. | GBP & 80.72 & \textbf{75.07} & 68.68 & 62.49 & 48.33 & 42.2 & 36.33 & \textbf{31.66}         \\
                        Resnet-34 | Std. | GBP & 79.53 & 74.2 & 68.19 & 62.48 & 46.66 & 40.95 & 35.73 & 30.83 \\
                        Resnet-34 | Adv. | GBP & 80.15 & 74.60 & 68.47 & 62.53  & 47.38 & 42.0 & \textbf{36.34} & 31.53     \\
                        Resnet-10 | Adv. | G.CAM++ & 79.67 & 74.05 & 68.12 & 61.84  & 47.28 & 41.72 & 35.99 & 31.05      \\ 
                        Resnet-34 | Adv. | G.CAM++  &79.74 & 74.5 & \textbf{68.87} & \textbf{62.68} & 46.12 & 40.47 & 35.31 & 30.81\\ 
\hline 
& & & & & & & & \\ 
\hline
                          TRADES & \textbf{84.0} & 73.25 &	59.79 &	47.03 & 47.21 & 42.2 & 37.03 & 32.96 \\
                          TRADES + Uniform-Noise & 81.69 & 74.96 &67.43 & 60.05 & \textbf{51.90} & 42.85 & 37.30 & 31.71\\
                          \cline{1-9}
                          \textbf{TRADES-SAT} & & & & \\ \cline{1-1}
                        Resnet-10 | Std. | GBP & 80.15 & 75.2 & 69.0 & 63.2 & 48.63 & 42.91 & 37.79 & 33.19 \\
                          Resnet-10 | Adv. | GBP & 80.65 & 75.38 & 69.28 & \textbf{63.46}  & 48.74 & 42.99 & 37.83 & 33.23 \\
                          Resnet-34 | Std. | GBP & 79.98 & 74.5 & 68.56 & 62.43 & 48.5 & 42.41 & 36.93 & 32.01 \\
                          Resnet-34 | Adv. | GBP & 80.26 & 74.87 & 68.75 & 62.74   & 48.76 & 42.83 & 37.36 & 32.87   \\
                          Resnet-10 | Adv. | G.CAM++ & 79.85 & 74.61 & 69.07 & 63.2 & 48.99 & 43.05 & 37.4 & 32.84      \\ 
                         Resnet-34 | Adv. | G.CAM++  & 83.17 & \textbf{77.18} & \textbf{70.27}	& 62.87 & 49.35 & \textbf{43.62} & \textbf{38.53} & \textbf{33.93}
                                  
                           \\ \hline 

\end{tabular}

}
\end{center}

\caption{Results on CIFAR-10 and CIFAR-100 using saliency maps from different teachers and explanation methods. $\epsilon \in \{1/255,2/255,3/255,4/255\}$ denotes the maximum $l_\infty$ perturbation allowed in 5-step PGD attack (More the $\epsilon$, stronger the attack). GBP:  Guided-Backpropagation; G.CAM++: Grad-CAM++.}
\label{table:cifar-10}
\end{table}

\noindent {\large \textbf{4.2 Results}}

\noindent Tables \ref{table:cifar-10} and \ref{table:bbox} present our results on CIFAR-10 + CIFAR-100 and Tiny ImageNet + Flower-17 datasets respectively. We note again that the provided bounding boxes are used as saliency maps in Tiny ImageNet, and segmentation masks provided in Flower-17 are used as saliency maps. We note that with no additional cost of computing perturbations, we obtain improvement of $~2-4\%$ in robustness. Table \ref{table:cifar-10} shows the results for CIFAR-10 and CIFAR-100, and demonstrates the potential of using saliency maps for adversarial training. Barring 1-2 cases, the proposed SAT (or its combination with an adversarial training method such as PGD or TRADES) performs better than respective baselines across the results. As the adversarial attack gets stronger (larger $\epsilon$), adding SAT to existing methods significantly improves robustness performance (~1-3\% in adversarial accuracy).
\begin{table}
\begin{center}
\scalebox{0.9}{
\footnotesize
\begin{tabular}{c|c|c|c|c|c|c}
\hline
\textbf{Method}  & \multicolumn{3}{c|}{\textbf{Tiny-Imagenet}} & \multicolumn{3}{c}{\textbf{FLOWER-17}} \\
 & $\epsilon=1/255$ & $\epsilon=2/255$  & $ \epsilon=3/255$  & $\epsilon=1/255$ & $\epsilon=2/255$  & $ \epsilon=3/255$ \\
\hline \hline
                        Original & 1.04 & 0.4 & 0.0  & 63.2 & 48.01 & 34.2  \\ 
                        Original + Uniform-Noise & 9.45 & 2.32 & 0.77 & 64.56 & 50.43 & 36.2 \\
                        SAT & \textbf{9.79}  & \textbf{2.46} & \textbf{0.77} & \textbf{66.17}  &\textbf{ 52.94} & \textbf{38.93} \\
                        \cline{1-7}
                        PGD & 18.91 & 14.34 &11.37 & 72.38 & 70.4 & 70.3 \\ 
                        PGD + Uniform-Noise  & 19.57 & 15.49 & 11.66 & 73.52 & 72.79 & 72.71 \\ 

                        PGD-SAT  & \textbf{20.56} & \textbf{16.38} & \textbf{12.91} & \textbf{78.67} & \textbf{75.73} & \textbf{75.00} \\
                        \cline{1-7}
                        TRADES & 18.45 & 16.76 &	11.09& 74.56 & 73.89 &	73.67 \\
                        TRADES + Uniform-Noise & 19.96 & 16.13 & 12.58 & 76.47 & 74.26 & 74.0 \\
                        TRADES-SAT& \textbf{20.04} & \textbf{16.45} & \textbf{12.96} & \textbf{79.41} & \textbf{77.94} & \textbf{77.20} \\
                        
\hline
\end{tabular}}
\end{center}
\caption{Results on Tiny-Imagenet and Flower-17 datasets where bounding boxes and segmentation masks provided with the dataset are used as saliency maps, respectively. $\epsilon \in \{1/255,2/255,3/255\}$ denotes the maximum $l_\infty$ perturbation allowed in 5-step PGD attack (More the $\epsilon$, stronger the attack).}
\vspace{-25pt}
\label{table:bbox}
\end{table}
It is also evident from Table \ref{table:cifar-10} that using finer saliency maps (GBP in our results) obtains better performance than weaker saliency maps (bounding boxes obtained from GradCAM++, in our case). This supports the inference that a stronger saliency map provides better adversarial robustness. We also notice, as pointed out earlier in this section (Fig \ref{saliency}), that using saliency maps obtained from an adversarially trained teacher leads to significant gains in robustness/adversarial accuracy performance. In other words, all our experiments point to the inference that better the saliency maps, better the adversarial robustness of the model trained using our approach. This further supports our inherent claim that saliency maps do provide adversarial robustness. 

Table \ref{table:cifar-10} also shows that PGD-SAT and TRADES-SAT leads to improvement ($~1.5 - 3\%$) over vanilla PGD and TRADES adversarial training. This shows that adding our saliency-based method to existing adversarial training method further improves their performance. We observe here again that using saliency maps of an adversarially robust teacher model proves more useful than those obtained from a normal teacher model. 

\begin{wrapfigure}[14]{r}{0.4\textwidth}
\includegraphics[width=0.4\textwidth]{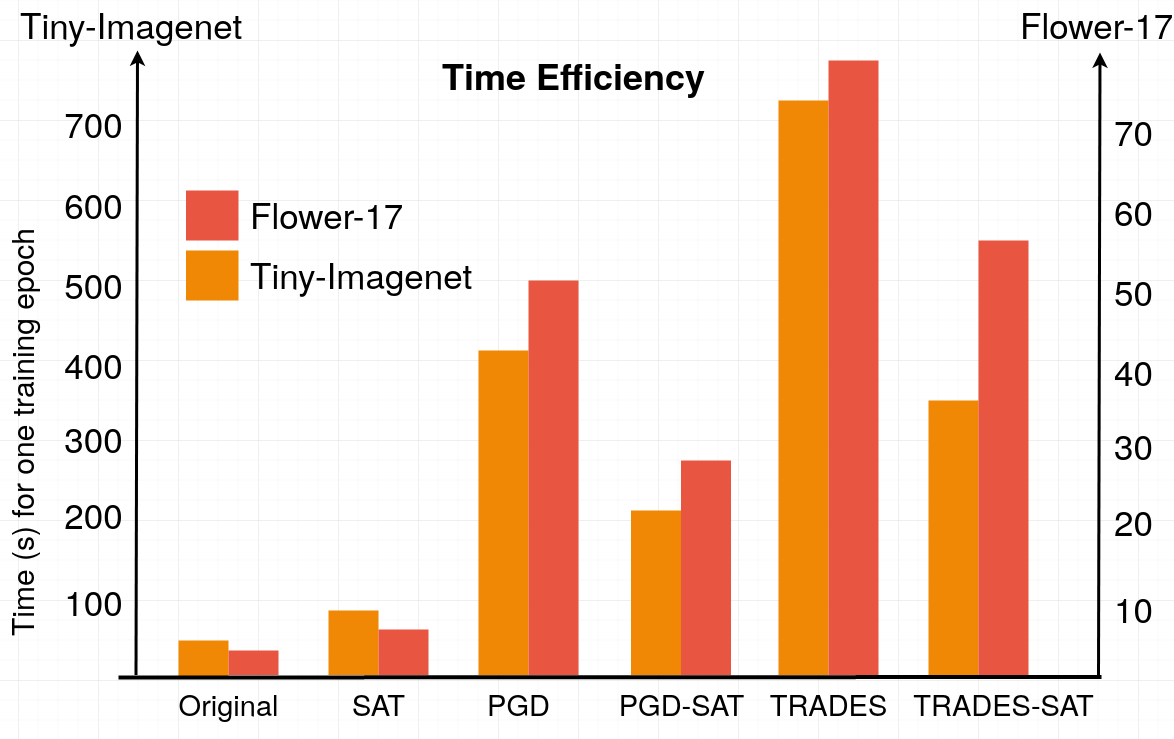}
\caption{Training time for one epoch in seconds (averaged over 10 trials) for different methods considered in our results on Tiny-Imagenet and Flower-17.}
\label{fig:train-time}
\end{wrapfigure}
\noindent \textbf{Time Efficiency:}
We further analyzed the training efficiency of the proposed method. Figure \ref{fig:train-time} reports the average time taken by one epoch over 10 trials. As can be seen, PGD-SAT and TRADES-SAT require only $\approx 50-70\%$ of training time when compared to vanilla PGD and TRADES respectively, and at the same time, achieves superior performance (as in Table \ref{table:bbox}). In case of vanilla SAT, the behavior is desirable since we observe an increase in robustness without compromising much in training time. We assume in these results that explanations are provided to us, inspiring the demand for finer saliency map annotations to be included in vision datasets to the community. Similar results for CIFAR-10 and CIFAR-100 are included in the appendix.



\section{Discussions, Ablation Studies and Conclusions}
\label{ablations}
We carried out ablation studies to better characterize the efficacy of our methodology, and present these results in this section. Unless explicitly specified, these experiments are carried out on CIFAR-100 using GuidedBackprop and GradCAM++ explanations of a standard and adversarially trained Resnet-10 teacher.\\

\begin{wrapfigure}[11]{l}{0.5\textwidth}
\vspace{-20pt}
\includegraphics[width=0.5\textwidth]{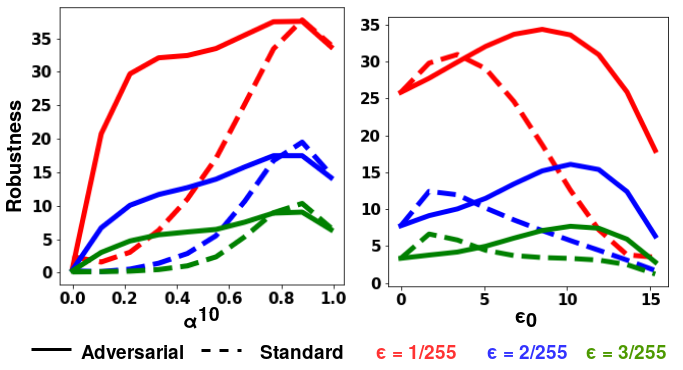}
\vspace{-20pt}
\caption{Variation of hyper-parameter $\alpha$ and $\epsilon_0$ on CIFAR-100 dataset.}
\label{fig:variation-alpha}
\end{wrapfigure}

      

\noindent \textbf{Varying Hyperparameters:} We studied the effect on the adversarial robustness of our trained model by varying the hyperparameters $\alpha$ and $\epsilon_0$ in our method. Fig \ref{fig:variation-alpha} shows these results, where all models are evaluated using a 5-Step PGD attack with $l_\infty$ norm and maximum perturbation $\epsilon \in \{\frac{1}{255}, \frac{2}{255}, \frac{3}{255} \}$.
We achieve higher robustness when $\alpha^{10}$ is set closer to 1 ($\alpha$ close to 1), which from Eqn \ref{pert-calc} indicates that the saliency map is useful to obtain perturbations in later stages of training, as explained before. Interestingly, when $\alpha^{10}$ is close to  0 ($\alpha$ close to 0), the training doesn't include any noise factor, and since we have a fixed saliency map for each image, it becomes equivalent to training on an additively shifted version of the original training data resulting in a less robust model.
While varying $\epsilon_0$ in Algorithm \ref{mimic-AT}, we observe a peak in robustness somewhere in the middle of the considered range of values. In Algorithm \ref{mimic-AT}, the distribution of our estimated perturbations and actual adversarial perturbations depend on the hyperparameter $\epsilon_0$. At $\epsilon_0 = 0$, the distributions are identical, and the distributions diverge as $\epsilon_0$ increases. Hence, at high values of $\epsilon_0$, when both distributions diverge, the perturbations used by SAT no longer resemble true adversarial perturbations, hence resulting in less robust models (when attacked adversarially using methods associated with the adversarial perturbations). In other words, the perturbations provided by saliency maps in our method has reasonable conjunction with standard adversarial perturbations, as shown in Fig \ref{fig:sal-attack}.


\begin{wraptable}[16]{l}{0.5\textwidth}
\vspace{-20pt}
\scalebox{0.65}{
\begin{tabular}{c|c|c|c|c}
\hline
\textbf{Method}  & \multicolumn{4}{c}{\textbf{PGD}}  \\
 & $\frac{1}{255}$ & $\frac{2}{255}$ & $\frac{3}{255}$ & $\frac{4}{255}$  \\
& & & &  \\
\hline \hline
Original  & 25.83 & 7.76 & 3.35 & 1.94 \\
Original + Uniform-Noise & 33.15 & 13.50 & 6.01 & 3.22 \\
\cline{1-5}
\textbf{SAT} & & & & \\ \cline{1-1}
Resnet-10 | Std. | GBP  & 20.53 & 7.52 & 3.5 & 2.12  \\
Resnet-10 | Std. | S.Grad  & \textbf{39.22} & \textbf{19.89} & \textbf{9.44} & \textbf{4.49}  \\
Resnet-10 | Std. | G.G.CAM++ & 21.46 & 8.00 & 3.53 & 2.23  \\
Resnet-10 | Std. | I.Grad & 36.2 & 5.43 & 7.28 & 3.37  \\
                           
Resnet-10 | Adv. | GBP  & 34.29 & 14.73 & 6.84 & 4.22  \\
Resnet-10 | Adv. | S.Grad  & \textbf{40.01} & \textbf{21.2} & \textbf{10.96} & \textbf{4.85}  \\
Resnet-10 | Adv. | G.G.CAM++ & 34.07 & 13.18 & 5.85 & 3.09  \\
Resnet-10 | Adv. | I.Grad & 37.56 & 16.45 & 7.55 & 4.31 \\
                           \hline
\end{tabular}}
\caption{\footnotesize Results using saliency maps from Guided-Backprop (GBP), Smooth-Grad (S.Grad), Integrated-Gradients (I.Grad), and Guided-Grad-CAM++ (G.G.CAM++)}
\label{table:better-exp}
\end{wraptable}

\noindent \textbf{Robustness against Other Attacks:}
In Table \ref{table:cifar-10}, we evaluate our models using the widely used PGD attack. We now study how the proposed method works against other attacks - in particular, Uniform Noise attack, TRADES \cite{trades} attack and a Saliency-based adversarial attack, which is an attack using saliency maps generated by our method. 
These results are shown in Table \ref{table:diff-pert}. It is clear that SAT and PGD-SAT outperform or have comparable performance to (in case of Uniform Noise) competing methods. Expectedly, the improvement is much more significant in case of Saliency-based adversarial attack, since the method is closely associated with the attack. 

\begin{table}
\centering
\begin{center}
\scalebox{0.8}{
\footnotesize
\begin{tabular}{c|c|c|c|c|c|c|c|c|c}
\hline

\textbf{Method} &  \multicolumn{3}{c|}{\textbf{Uniform-Noise}} & \multicolumn{3}{c|}{\textbf{TRADES}} &
      \multicolumn{3}{c}{\textbf{Saliency Attack}} \\
         & $\epsilon=\frac{8}{255}$ & $\epsilon=\frac{16}{255}$ & $\epsilon=\frac{32}{255}$ & $\epsilon=\frac{2}{255}$ & $\epsilon=\frac{3}{255}$ & $\epsilon=\frac{4}{255}$  & $\epsilon=\frac{4}{255}$ & $\epsilon=\frac{8}{255}$ & $\epsilon=\frac{16}{255}$  \\
        &  &  &  & &  &  &  &  &  \\
        
\hline \hline
Original & 62.21 & 47.88 & 21.17 &  39.59
                            & 35.1 & 33.45  & 13.03 & 6.86 & 3.92
                            \\
Original + Uniform-Noise  & \textbf{67.42} & \textbf{60.75} & 28.9 &  \textbf{46.98}
                            & 39.37 & 37.8 & 19.63 & 10.48 & 6.1
                            \\
\hline
\textbf{SAT} &  & &  &  & & & & & \\ \cline{1-1}
Resnet-10 | Std | GBP & 56.63 & 57.4 & \textbf{46.9} & 36.45 & 33.35 & 32.46  & 51.78 & 49.02 & 42.75             \\

Resnet-10 | Adv. | GBP  & 66.63 & 57.93 & 21.92  & 45.93 & \textbf{40.80} & \textbf{37.86} &  62.99 & 56.48 & 43.49              \\

Resnet-10 | Adv. | G.CAM++  & 64.00 & 54.84 & 23.43 & 44.31 & 38.28 & 35.71 & \textbf{63.86} & \textbf{63.87} & \textbf{63.66}               \\
                           \hline \hline
PGD  & 50.96 & 51.0 & 48.69  & 49.03 & 47.63 & 45.34 & 44.03 & 34.77  & 21.04             \\
PGD + Uniform-Noise   & \textbf{55.6} & 52.45 & 49.20 & \textbf{52.32}
                            & 48.76 & 45.87  & 45.72 & 32.81 & 16.62
                            \\
\hline
\textbf{PGD-SAT} &  & &  &  & & & & &  \\ \cline{1-1}
Resnet-10 | Std | G.BP  & 52.78 & 52.56 & 49.21  & 50.43 & 47.69 & 46.24 &  47.25 & 46.74 & 44.95             \\
Resnet-10 | Adv. | G.BP   & 54.38 & \textbf{54.13} & \textbf{50.30}  & 51.58 & \textbf{49.15} & \textbf{46.83}   & 50.61 & 50   & 47.64            \\
Resnet-10 | Adv. | G.CAM++  & 53.30 & 52.76 & 49.74  & 50.87 & 48.42 & 46.31  &  \textbf{50.69} & \textbf{50.63}  & \textbf{50.54}            \\
                           \hline                           
\end{tabular}
}
\end{center}
\caption{Results against other attacks: Uniform-Noise, TRADES and Saliency. \textit{(GBP=  Guided-Backprop; G.CAM++=Grad-CAM++; $l_\infty $ perturbation $\in [-\epsilon, \epsilon]$.)}}
\vspace{-25pt}
\label{table:diff-pert}
\end{table}

\noindent \textbf{Using Other Saliency Maps:}
We also performed a study where we trained our models using stronger saliency maps obtained using methods such as SmoothGrad \cite{smooth-grad}, Guided Grad-CAM++ \cite{grad-cam++} and Integrated Gradients \cite{integrated-gradients} from teacher models. These models are evaluated using a 5-Step PGD attack with $l_\infty$ max-perturbation $\epsilon \in \{\frac{1}{255}, \frac{2}{255}, \frac{3}{255}, \frac{4}{255}\}$. As can be seen from Table \ref{table:better-exp}, we tend to achieve more robust models for better saliency maps.\\ 

\noindent \textbf{Conclusion.} In summary, this work explores the interesting connection between saliency maps and adversarial robustness to propose a Saliency based Adversarial training (SAT) method. SAT imitates adversarial training by using saliency maps to mimic adversarial perturbations. In particular, our methodology allows the use of annotations such as bounding boxes and segmentation masks to be exploited as weak explanations to improve model's robustness with no additional computations required to compute the perturbations themselves. Our results on CIFAR-10, CIFAR-100, Tiny ImageNet and Flowers-17 corroborate our claim. We further gain improvement over popular adversarial training methods by integrating SAT with them (PGD-SAT and TRADES-SAT). Further, our work shows how using better saliency maps leads to more robust models. Our effort opens rather a new direction to enhance robustness of DNNs by exploiting saliency maps, and inspires the need for strong saliency maps to be provided with vision datasets, which helps train adversarially robust models with little overhead. Our future work includes ways to improve SAT by reasoning about the class closest to the ground truth in terms of decision boundary, and improving our estimates of adversarial perturbations. 

\section*{Acknowledgement}
\vspace{-3pt}
We are grateful to the Ministry of Human Resource Development, India; Department of Science and Technology, India; as well as Honeywell India for the financial support of this project through the UAY program. We thank the anonymous reviewers for their valuable feedback that helped improve the presentation of this work.

%
%
%

{\bibliographystyle{splncs04} \bibliography{egbib}}

\begin{thebibliography}{10}
\providecommand{\url}[1]{\texttt{#1}}
\providecommand{\urlprefix}{URL }
\providecommand{\doi}[1]{https://doi.org/#1}

\bibitem{var-grad}
Adebayo, J., Gilmer, J., Goodfellow, I.J., Kim, B.: Local explanation methods
  for deep neural networks lack sensitivity to parameter values. CoRR  (2018)

\bibitem{jacobianAT}
Chan, A., Tay, Y., Ong, Y.S., Fu, J.: Jacobian adversarially regularized
  networks for robustness. In: ICLR'20

\bibitem{grad-cam++}
Chattopadhyay, A., Sarkar, A., Howlader, P., Balasubramanian, V.N.: Grad-cam++:
  Generalized gradient-based visual explanations for deep convolutional
  networks. In: WACV'18

\bibitem{attribute-regularization}
Chen, J., Wu, X., Rastogi, V., Liang, Y., Jha, S.: Robust attribution
  regularization. In: NeuRIPS'19

\bibitem{parseval-nets}
Cisse, M., Bojanowski, P., Grave, E., Dauphin, Y., Usunier, N.: Parseval
  networks: Improving robustness to adversarial examples. In: ICML'17

\bibitem{tiny-img}
CS231N, S.: Tiny ImageNet Visual Recognition Challenge,
  \url{https://tiny-imagenet.herokuapp.com/}

\bibitem{explanations-manipulated}
Dombrowski, A.K., Alber, M., Anders, C.J., Ackermann, M., M\"{u}ller, K.R.,
  Kessel, P.: Explanations can be manipulated and geometry is to blame. In:
  NeuRIPS'19

\bibitem{connection@etmann}
Etmann, C., Lunz, S., Maass, P., Schönlieb, C.B.: On the connection between
  adversarial robustness and saliency map interpretability. In: ICML'19

\bibitem{imagenet-texture}
Geirhos, R., Rubisch, P., Michaelis, C., Bethge, M., Wichmann, F.A., Brendel,
  W.: Imagenet-trained cnns are biased towards texture; increasing shape bias
  improves accuracy and robustness. In: ICLR'19

\bibitem{fragile-interpretation}
Ghorbani, A., Abid, A., Zou, J.: Interpretation of neural networks is fragile.
  In: AAAI'19

\bibitem{at@goodfellow}
Goodfellow, I., Shlens, J., Szegedy, C.: Explaining and harnessing adversarial
  examples. In: ICLR'15 (2015)

\bibitem{relating-explanations}
Ignatiev, A., Narodytska, N., Marques-Silva, J.: On relating explanations and
  adversarial examples. In: NeuRIPS'19

\bibitem{patternnet}
Kindermans, P.J., Schütt, K.T., Alber, M., Müller, K.R., Erhan, D., Kim, B.,
  Dähne, S.: Learning how to explain neural networks: Patternnet and
  patternattribution. In: ICLR'18

\bibitem{cifar-datasets}
Krizhevsky, A.: Learning multiple layers of features from tiny images (2009)

\bibitem{lrp}
Lapuschkin, S., Binder, A., Montavon, G., Klauschen, F., Müller, K.R., Samek,
  W.: On pixel-wise explanations for non-linear classifier decisions by
  layer-wise relevance propagation  (2015)

\bibitem{at@madry}
Madry, A., Makelov, A., Schmidt, L., Tsipras, D., Vladu, A.: Towards deep
  learning models resistant to adversarial attacks. In: ICLR'18 (2018)

\bibitem{flower}
{Nilsback}, M.., {Zisserman}, A.: A visual vocabulary for flower
  classification. In: CVPR'06

\bibitem{pertexp1}
Ribeiro, M.T., Singh, S., Guestrin, C.: “why should i trust you?”:
  Explaining the predictions of any classifier. In: ACM SIGKDD'16

\bibitem{defense-gan}
Samangouei, P., Kabkab, M., Chellappa, R.: Defense-{GAN}: Protecting
  classifiers against adversarial attacks using generative models. In: ICLR'18

\bibitem{grad-cam}
Selvaraju, R.R., Das, A., Vedantam, R., Cogswell, M., Parikh, D., Batra, D.:
  Grad-cam: Why did you say that? visual explanations from deep networks via
  gradient-based localization. In: ICCV'17

\bibitem{free-AT}
Shafahi, A., Najibi, M., Ghiasi, A., Xu, Z., Dickerson, J.P., Studer, C.,
  Davis, L.S., Taylor, G., Goldstein, T.: Adversarial training for free! In:
  NeuRIPS'19

\bibitem{deep-lift}
Shrikumar, A., Greenside, P., Kundaje, A.: Learning important features through
  propagating activation differences. In: ICML'17 (2017)

\bibitem{LAT}
Sinha, A., Singh, M., Kumari, N., Krishnamurthy, B., Machiraju, H.,
  Balasubramanian, V.N.: Harnessing the vulnerability of latent layers in
  adversarially trained models. In: IJCAI'19

\bibitem{smooth-grad}
Smilkov, D., Thorat, N., Kim, B., Vi{\'{e}}gas, F.B., Wattenberg, M.:
  Smoothgrad: removing noise by adding noise. CoRR  (2017)

\bibitem{gbp}
Springenberg, J., Dosovitskiy, A., Brox, T., Riedmiller, M.: Striving for
  simplicity: The all convolutional net. In: ICLR (workshop track) (2015)

\bibitem{integrated-gradients}
Sundararajan, M., Taly, A., Yan, Q.: Axiomatic attribution for deep networks.
  In: ICML'17

\bibitem{szegedy2013intriguing}
Szegedy, C., Zaremba, W., Sutskever, I., Bruna, J., Erhan, D., Goodfellow, I.,
  Fergus, R.: Intriguing properties of neural networks. In: ICLR'14

\bibitem{robustness-odds}
Tsipras, D., Santurkar, S., Engstrom, L., Turner, A., Madry, A.: Robustness may
  be at odds with accuracy. In: ICLR'19

\bibitem{fgvis}
Wagner, J., Mathias~Köhler, J., Gindele, T., Hetzel, L., Thaddäus~Wiedemer,
  J., Behnke, S.: Interpretable and fine-grained visual explanations for
  convolutional neural networks. In: CVPR'18

\bibitem{spatial-adv-exmp}
Xiao, C., Zhu, J.Y., Li, B., He, W., Liu, M., Song, D.: Spatially transformed
  adversarial examples. In: ICLR'18

\bibitem{feature-denoising}
Xie, C., Wu, Y., van~der Maaten, L., Yuille, A.L., He, K.: Feature denoising
  for improving adversarial robustness. In: CVPR'19

\bibitem{pertexp2}
Zeiler, M.D., Fergus, R.: Visualizing and understanding convolutional networks

\bibitem{YOPO}
Zhang, D., Tianyuan, Z., Lu, Y., Zhu, Z., Dong, B.: You only propagate once:
  Painless adversarial training using maximal principle. In: NeuRIPS'19

\bibitem{trades}
Zhang, H., Yu, Y., Jiao, J., Xing, E.P., Ghaoui, L.E., Jordan, M.I.:
  Theoretically principled trade-off between robustness and accuracy. In:
  ICML'19

\bibitem{excitation-bp}
Zhang, J., Lin, Z., Brandt, Jonathan, S.X., Sclaroff, S.: Top-down neural
  attention by excitation backprop. In: ECCV'16

\bibitem{interpret-atcnn}
Zhang, T., Zhu, Z.: Interpreting adversarially trained convolutional neural
  networks. In: ICML'18

\bibitem{cam}
Zhou, B., Khosla, A., Lapedriza, {\`{A}}., Oliva, A., Torralba, A.: Learning
  deep features for discriminative localization. In: CVPR'16

\end{thebibliography}

\section*{\huge Appendix}

In this appendix, we include additional results that could not be included in the main paper due to space constraints.\\

\noindent \textbf{Dataset details:} \textit{CIFAR-10} is a subset of 80 million tiny images dataset and consists of 60,000 $32\times32$ color images containing one of 10 object classes, with 6000 images per class. \textit{CIFAR-100} is just like CIFAR-10, except that it has 100 classes containing 600 images each. There are 500 training images and 100 testing images per class. \textit{Tiny Imagenet} has 200 classes, with each class containing 500 training images, 50 validation images, and 50 test images. The images are of resolution 64 x 64. \textit{FLOWER-17} is a 17 category flower dataset with 80 images for each class. The flowers chosen are some common flowers in the UK. The images have large scale, pose and light variations and there are also classes with large variations of images within the class and close similarity to other classes.

\noindent \textbf{Time Efficiency:} In Figure \ref{fig:train-time}, we report the training efficiency of proposed methods for CIFAR-10 and CIFAR-100 datasets. As evident from the figure, we achieve analogous results as Tiny-Imagenet and Flower-17 (see Fig 4 of main submission).

\begin{figure}
\centering
\includegraphics[width=0.7\textwidth]{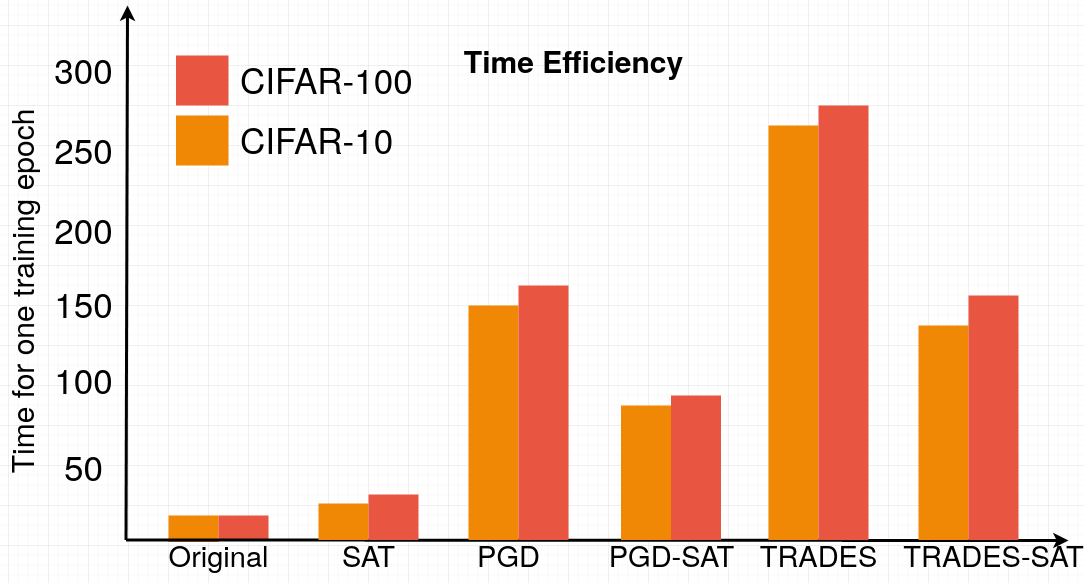}
\caption{Training time for one epoch in seconds (averaged over 10 trials) for different methods considered in our results on CIFAR-10/-100.}
\label{fig:train-time}
\end{figure}

\noindent \textbf{Robustness to Stronger PGD Attacks:}
As shown in the main paper (Section 4), our trained student models are robust against the 5-Step PGD [16] attack. We also studied the effect of even stronger attacks on our trained student models. The models are evaluated using 20-Step and 40-Step PGD attacks with $l_\infty$ max-perturbation $\epsilon \in \{\frac{1}{255}, \frac{2}{255}, \frac{3}{255}, \frac{4}{255}\}$. These results are shown in Table \ref{table:strong-attack-robust}. As the results show, our models show similar improvements in robustness even on these stronger attacks.\\

\begin{table}
\begin{center}
\scalebox{0.9}{
\begin{tabular}{c|c|c|c|c|c}
\hline
\textbf{Method} &  \textbf{Attack} &$ \epsilon = 1/255$ & $ \epsilon =2/255$ & $\epsilon = 3/255$ & $\epsilon = 4/255$ \\
& & & & & \\
\hline \hline
\multirow{2}{*}{Original} 
                            & PGD-20 & 27.17 & 6.94 & 2.24 & 0.81             \\
                            & PGD-40 & 27.08 & 6.86 & 2.06 & 0.72             \\
                            \cline{1-6}
\multirow{2}{*}{Original + Uniform-Noise} 
                            & PGD-20 & \textbf{36.39} & 12.22 & 5.25 & 2.29            \\
                            & PGD-40 & \textbf{36.45} & 12.1 & 5.14 & 2.18            \\
                            \cline{1-6}
\textbf{SAT} & & & &  &\\ \cline{1-1}
\multirow{2}{*}{Resnet-10 | Std | GBP} 
                            & PGD-20 & 20.87 & 7.25 & 2.76 & 1.21            \\
                            & PGD-40 & 20.92 & 7.18 & 2.6 & 1.19             \\
                            \cline{1-6}

\multirow{2}{*}{Resnet-10 | Adv. | GBP} 
                           & PGD-20 & 34.79 & \textbf{13.94} & \textbf{5.28} & \textbf{2.38}             \\
                            & PGD-40 & 34.97 & \textbf{13.97} & \textbf{5.2} & \textbf{2.27}             \\
                            \cline{1-6}

\multirow{2}{*}{Resnet-10 |Adv. | G.CAM++} 
                            & PGD-20 & 33.46 & 12.55 & 5.06 & 2.37             \\
                            & PGD-40 & 33.52 & 12.46 & 4.98 & 2.27             \\
\hline \hline 
\multirow{2}{*}{PGD} 
                            & PGD-20 & 45.75 & 40.09 & 35.19 & 30.67             \\
                            & PGD-40 & 45.76 & 40.09 & 35.18 & 30.65             \\
                            \cline{1-6}
\multirow{2}{*}{PGD + Uniform-Noise} 
                            & PGD-20 & \textbf{50.76} & 41.61 & 35.05 & 30.58            \\
                            & PGD-40 & \textbf{50.75} & 41.63 & 35.03 & 30.52            \\
                            \cline{1-6}
\textbf{PGD-SAT} & & & &  &\\ \cline{1-1}
\multirow{2}{*}{Resnet-10 | Std | GBP} 
                                & PGD-20 & 44.89 & 39 & 34.14 & 29.23             \\
                             & PGD-40 & 44.89 & 39.02 & 34.14 & 29.17             \\
                            \cline{1-6}

\multirow{2}{*}{Resnet-10 | Adv. | GBP} 
                            & PGD-20 & 48.10 & \textbf{42.13} & \textbf{36.16} & \textbf{31.1}            \\
                            & PGD-40 & 48.11 & \textbf{42.14} & \textbf{36.14} & \textbf{31.07}             \\
                            \cline{1-6}
\multirow{2}{*}{Resnet-10 | Adv. | G.CAM++} 
                           & PGD-20 & 47.29 & 41.69 & 35.98 & 30.96             \\
                         & PGD-40 & 47.29 & 41.69 & 35.92 & 30.93             \\
                            \cline{1-6}
                          \hline                           
      
\end{tabular}
}
\end{center}
\caption{Improved robustness to stronger PGD attacks on CIFAR-100. GBP: Guided-Backpropogation; G.CAM+: Grad-CAM++.}
\label{table:strong-attack-robust}
\end{table}

\noindent \textbf{Using Saliency Maps of Sparse Teachers:}
We add another variation in the quality of saliency maps wherein we use the saliency maps of a sparse teacher model  (with 95\% of total weights as zero) whose classification performance is however comparable to a dense equivalent model. We observed that saliency maps of sparse teachers are slightly better than standard ones but not better than that of adversarially trained teachers and expect similar behavior when used in SAT. Therefore, we trained SAT and PGD-SAT models on CIFAR-10 using Guided-Backpropogation [25] of Resnet-10/Resnet-34 sparse teachers to provide the saliency map. Table \ref{table:sparse} shows our results. We notice that in most cases, using saliency maps of sparse teachers leads to reasonable improvement as compared to saliency maps of the standard teacher but short of adversarially trained teachers. 

\begin{table}
\begin{center}
\scalebox{1}{
\begin{tabular}{c|c|c|c|c}
\hline

\textbf{Method}  & $\epsilon = 1/255$ & $\epsilon = 2/255$  & $\epsilon = 3/255$ & $\epsilon = 4/255$ \\
& & & &  \\
\hline \hline
                        Original & 47.71 & 10.36 & 1.39 & 0.28 \\ 
                        Original + Uniform-Noise & 61.23 & 22.85 & 6.34 & 2.56 \\
                        \cline{1-5}
                        \textbf{SAT} & & & & \\ \cline{1-1}  
                        Resnet-10 | Std. | GBP & 10.87  & 0.95  & 0.0  & 0.0              \\
                         Resnet-10 | Adv. | GBP & \textbf{63.33}  & \textbf{26.79} & \textbf{9.62} & \textbf{3.69}          \\ 
                         Resnet-34 | Std. | GBP  & 18.01 & 2.54 & 1.25 & 1.0             \\
                         Resnet-34 | Adv. | GBP & 62.67 & 23.76  & 5.82 & 1.28 \\ 
                        Resnet-10 | Sparse | GBP & 52.43 & 26.79 & 5.82 & 1.97           \\
                        Resnet-34 | Sparse | GBP & 60.18& 23.56 & 8.58 & 2.76 \\
\hline \hline 
                        PGD & 77.89 & 73.1 & 66.96 & 61.18  \\ 
                        PGD + Uniform-Noise & \textbf{82.23} & 74.97 & 65.61 & 55.04 \\ 
                        \cline{1-5}
                        \textbf{PGD-SAT} & & & & \\
                        \cline{1-1}
                        Resnet-10 | Std. | GBP & 79.72 & 73.81 & 67.72 & 61.29   \\
                        Resnet-10 | Adv. | GBP & 80.72 & \textbf{75.07} & \textbf{68.68} & 62.49          \\
                        Resnet-34 | Std. | GBP & 79.53 & 74.2 & 68.19 & 62.48  \\
                        Resnet-34 | Adv. | GBP & 80.15 & 74.60 & 68.47 & \textbf{62.53}      \\
                         Resnet-10 | Sparse | GBP & 79.48 & 73.97 & 67.72 & 61.77          \\ 
                         Resnet-34 | Sparse| GBP  & 79.42 & 74.38 & 68.48 & 62.48 \\
 \hline

\end{tabular}
}

\end{center}
\caption{Results of improved robustness on CIFAR-10 using saliencies extracted from sparse teachers using guided-backpropagation. $\epsilon \in
\{1/255, 2/255, 3/255, 4/255\}$  denotes the maximum $l_\infty$ perturbation allowed in 5-step PGD attack (More the $\epsilon$, stronger the attack).}
\label{table:sparse}
\end{table}

\noindent \textbf{Integrating Multiple Saliency Maps}
In the main paper (Section 4), we show how integrating our estimated perturbations (SAT) with true adversarial perturbations leads to further improvement. It seems obvious then to question about the effect of using more than one saliency maps in SAT. To study this, we assess the contribution of using a combination of two saliency maps in SAT. For getting two different saliency maps, we either change: \textbf{(1)} model architecture (Resnet-10, Resnet-34) or  \textbf{(2)} training procedure (Sparse, Adv). We therefore select the saliency map to be used in SAT (Algorithm 1 in main paper) randomly between these two maps. We report our findings in Table \ref{table:integrate} (Shown using '+' symbol) for CIFAR-10. We observe that using a combination of saliency maps leads to improvement in some cases. The cases where the improvement is more, include saliency maps which have different characteristics. This can be attributed to the fact that coming from slightly two different distributions, the saliency maps now cover different types of adversarial perturbations than their singular equivalents. Consequently, one can infer that if richer saliency maps, perhaps from different sources of reasonable divergence are available, one can leverage them to improve performance further.

\begin{table}
\begin{center}
\scalebox{0.75}{
\begin{tabular}{c|c|c|c|c}
\hline

\textbf{Method}  & $\epsilon = 1/255$ & $\epsilon = 2/255$  & $\epsilon = 3/255$ & $\epsilon = 4/255$ \\
& & & &  \\
\hline \hline
                        Original & 47.71 & 10.36 & 1.39 & 0.28 \\ 
                        Original + Uniform-Noise & 61.23 & 22.85 & 6.34 & 2.56 \\
                        \cline{1-5}
                        \textbf{SAT} & & & & \\ \cline{1-1}  
                        Resnet-10 | Std. | GBP & 10.87  & 0.95  & 0.0  & 0.0              \\
                         Resnet-10 | Adv. | GBP & \textbf{63.33}  & \textbf{26.79} & \textbf{9.62} & \textbf{3.69}          \\ 
                         Resnet-34 | Std. | GBP  & 18.01 & 2.54 & 1.25 & 1.0             \\
                         Resnet-34 | Adv. | GBP & 62.67 & 23.76  & 5.82 & 1.28 \\ 
                          Resnet-10 | Adv. | GBP + Resnet-34 | Adv. | GBP & 
                          63.2  & 24.2 & 6.58 & 1.71             \\
                          Resnet-10 | Adv. | GBP + Resnet-34 | Sparse | GBP & \textbf{71.01} & \textbf{39.94} & \textbf{19.0} & \textbf{8.64}           \\
                          Resnet-10 | Sparse | GBP + Resnet-34 | Adv. | GBP  & 
                          66.47 & 35.19 & 15.0 & 6.11          \\ 
                          Resnet-10 | Sparse| GBP  + Resnet-34 | Sparse | GBP & 60.07 & 29.11 & 11.26 & 4.55 \\
 \hline

\end{tabular}
}

\end{center}
\caption{Results of improved robustness on CIFAR-10 by integrating saliencies extracted from different teachers and explanation methods. $\epsilon \in
\{1/255, 2/255, 3/255, 4/255\}$  denotes the maximum $l_\infty$ perturbation allowed in 5-step PGD attack (More the $\epsilon$, stronger the attack). '+' denotes which two saliencies are integrated together.}
\label{table:integrate}
\end{table}

\end{document}